\documentclass[final]{cvpr}

\usepackage{times}
\usepackage{epsfig}
\usepackage{graphicx}
\usepackage{amsmath}
\usepackage{amssymb}

\usepackage[pagebackref=true,breaklinks=true,colorlinks,bookmarks=false]{hyperref}

\def\cvprPaperID{****} % *** Enter the CVPR Paper ID here
\def\confYear{CVPR 2021}

\begin{document}

%%%%%%%%% TITLE
\title{Transfer Learning with Point Transformers}

\author{Kartik Gupta\\
{\small University of Massachusetts Amherst}\\
{\tt\small kgupta@umass.edu}
\and
Rahul Vippala\\
{\small University of Massachusetts Amherst}\\
{\tt\small rvippala@umass.edu}
\and
Sahima Srivastava\\
{\small University of Massachusetts Amherst}\\
{\tt\small sahimasrivas@umass.edu}
}

\maketitle

%%%%%%%%% ABSTRACT
\begin{abstract}
   Point Transformers are near state-of-the-art models for classification, segmentation, and detection tasks on Point Cloud data. They utilize a self attention based mechanism to model large range spatial dependencies between multiple point sets. In this project we explore two things: classification performance of these attention based networks on ModelNet10 dataset and then, we use the trained model to classify 3D MNIST dataset after finetuning. We also train the model from scratch on 3D MNIST dataset to compare the performance of finetuned and from-scratch model on the MNIST dataset. We observe that since the two datasets have a large difference in the degree of the distributions, transfer learned models do not outperform the from-scratch models in this case. Although we do expect transfer learned models to converge faster since they already know the lower level edges, corners, etc features from the ModelNet10 dataset. 
\end{abstract}

%%%%%%%%% BODY TEXT
\section{Introduction}

Point clouds are generated using specialised devices, like laser scanners or depth cameras, that emit signals and measure the time it takes for them the signals to bounce back. This is then used to estimate the distance/depth of different parts of an object. A large collection of such points results in a point cloud.\\
Point clouds offer a comprehensive and precise depiction of the three-dimensional realm. They accurately capture the form, arrangement, and spatial connections of objects and surroundings, exhibiting a high level of fidelity. As a result, point clouds find utility in applications that demand intricate geometric data, including fields like robotics, autonomous navigation, virtual reality, and augmented reality.\\

Point clouds also aid in object recognition in the 3D space. By looking at the arrangement of points, it is possible to identify objects or certain features of the objects. Point cloud segmentation involves assigning semantic labels to individual points in a point cloud. PointNet \cite{qi2017pointnet} is one of the most commonly used point cloud classification network that uses multi-layer perceptrons and pooling layers to extract features. PointNet++ \cite{qi2017pointnet++} improves upon the previous architecture by introducing hierarchical groupings of points into local regions allowing PointNet to be applied at different scales. This enables the model to capture both local and contextual features thus improving the classification performance. Further, CNN-based architectures like DGCNN \cite{wang2019dynamic} use k-nearest neighbour graphs to correlate points and apply graph convolutions to extract features. Relation Shape CNN \cite{liu2019relation} captures the geometric relationship between points by considering the position and orientation of points with respect to their neighbours. PointCNN \cite{li2018pointcnn} addresses the problem of unordered and irregularly sampled point clouds by introducing a learnable permutation-invariant function. This weighting function allows features to be extracted in an order-agnostic manner and facilitate better classification.\\
Transformers are a type of deep learning models that use attention mechanism to process information in a structured way~\cite{transformer, attention}. Point Transformers \cite{zhao2021point} uses the attention mechanism to capture relationship between points. This model utilizes positional encodings to include spatial information and applies multi-head self-attention to capture how points in the point cloud interact with each other. The transformer family of models is particularly appropriate for point cloud processing because the self-attention operator, which is at the core of transformer networks, is in essence a set operator: it is invariant to permutation and cardinality of the input elements. The application of self-attention to 3D point clouds is therefore quite natural, since point clouds are essentially sets embedded in 3D space.
We show that transformer based networks are superior to the models discussed above in variety of tasks such as classification, segmentation etc. We focus on the Point Transformer v1 architecture which has vector attention. We train the models on ModelNet10 dataset and transformer based networks achieve good accuracy in this task. 
We further explore the concept of transfer learning where we train the network on ModelNet10 dataset and then finetune it on 3D MNIST dataset. We compare performance of this model compared to one which was trained from scratch on 3D MNIST dataset.

\section{Related Work}
Point clouds in 3D space lack a specific order and are spread out, resembling sets. When it comes to handling these point clouds using learning-based methods, there are three main categories: projection-based networks, voxel-based networks, and point-based networks.

% MEHNAT_________________________________________
PointNet \cite{qi2017pointnet} directly handles unprocessed point cloud data for the purposes of 3D object classification and segmentation. PointNet exhibits a remarkable ability to capture both local and global geometric features, all while accommodating point cloud inputs that lack any predefined order or alignment.

The work in \cite{qi2017pointnet++} builds upon PointNet to make a PointNet++. PointNet++ enhances the representation learning capability by applying a set of PointNet modules in a nested fashion, capturing multi-scale features for improved object classification and segmentation.

The authors \cite{zhou2018voxelnet} introduce VoxelNet, a comprehensive learning framework designed for 3D object detection. VoxelNet specifically works with point clouds that have been transformed into volumetric representations. By utilizing a combination of convolutional and fully connected layers, VoxelNet effectively detects objects within the point cloud data, surpassing previous methods and achieving superior performance on different benchmark datasets.

% MEHNAT_________________________________________

% Point-based networks are deep network structures specifically designed to handle point clouds directly, without projecting or quantizing them onto regular grids in 2D or 3D. These networks treat point clouds as sets embedded in continuous space. PointNet and PointNet++ are examples of such networks that employ permutation-invariant operators, like pointwise MLPs and pooling layers, to aggregate features across the point set. PointNet++ further incorporates a hierarchical spatial structure to enhance sensitivity to local geometric arrangements. Efficient sampling strategies have also been developed to improve the performance of these models.

Another set of approaches connect the points in the cloud to form a graph and perform message passing on this graph. Methods like DGCNN \cite{wang2019dynamic}, PointWeb , ECC, SPG, KCNet, and others utilize different techniques to conduct graph convolutions or leverage contextual relationships within the point set. Some methods explore continuous convolutions directly applied to the 3D point set without quantization. For example, PCCN represents convolutional kernels as MLPs, SpiderCNN defines kernel weights using polynomial functions, and Spherical CNN addresses 3D rotation equivariance.

The introduction of Transformer and self-attention models has brought significant advancements to machine translation, natural language processing, and recommendation systems~\cite{proactive, neuroseqret, transformer}. These developments have also influenced the utilization of self-attention networks in 2D image recognition. Scalar dot-product self-attention has been applied within local image patches. Meanwhile, Zhao et al \cite{zhao2021point}. have introduced a range of vector self-attention operators.

The concept of self-attention is particularly relevant in the context of this study because it inherently operates on sets. Positional information is incorporated as attributes of the elements, treating them as a set. Since 3D point clouds essentially consist of points with positional attributes, the self-attention mechanism appears to be well-suited for this type of data. Consequently, a Point Transformer layer is developed to apply self-attention specifically to 3D point clouds.

% Previous works on point cloud analysis have also explored attention mechanisms, such as those by [48, 21, 50, 17]. However, these approaches employ global attention on the entire point cloud, leading to computational inefficiency and impracticality for large-scale 3D scene understanding. They also use scalar dot-product attention, where the same aggregation weights are shared across different channels. In contrast, the proposed method applies local self-attention, enabling scalability to scenes with millions of points, and utilizes vector attention, which is shown to be crucial for achieving high accuracy. Furthermore, the study highlights the significance of appropriate position encoding in large-scale point cloud understanding, in contrast to prior approaches that neglected position information.

% Overall, the research demonstrates that well-designed self-attention networks can effectively handle large and complex 3D scenes, significantly advancing the state of the art in large-scale point cloud understanding.

\section{Dataset}
We train our network on ModelNet10 \cite{wu20153d}, which is an extensive dataset utilized to train and assess 3D deep learning models. It encompasses a wide range of 3D computer-aided design (CAD) models, spanning ten distinct object categories like chairs, tables, and lamps. This dataset offers a rich variety of objects, featuring diverse characteristics such as different shapes, sizes, and appearances. 
We apply transfer learning on a 3D MNIST model obtained from Kaggle \footnote{https://www.kaggle.com/datasets/daavoo/3d-mnist?resource=download} which is created using the MNIST 2D dataset.

\section{Methodology}

With the aim to accurately categorise point clouds without using CNN's and without making any assumptions about the datapoints, we opted for a transformer-based model that utilises vectors for the purpose of classification. We built upon the work of Zhao et al in \cite{zhao2021point}. The authors argue that 3D point clouds are nothing but sets of points in space and so self-attention, which highlights positional information of a set of elements, is a useful tool. They used three datasets to demonstrate the utility of point transformers in 3D deep learning: For 3D Semantic Segmentation: Stanford Large-Scale 3D Indoor Spaces (S3DIS) \cite{armeni20163d}; For 3D Shape Classification: ModelNet40 dataset \cite{yi2016scalable}; For Object Part Segmentation: ShapeNetPart \cite{wu20153d}.\\
We utilise the same model. The input data for the model consists of labelled point clouds. The output of the model is a prediction score for each possible class. The architecture of the self-attention transformer block shown below. \\

% _______________________________________________________________
\begin{figure}[hbt]
\begin{center}
   \includegraphics[width=0.8\linewidth]{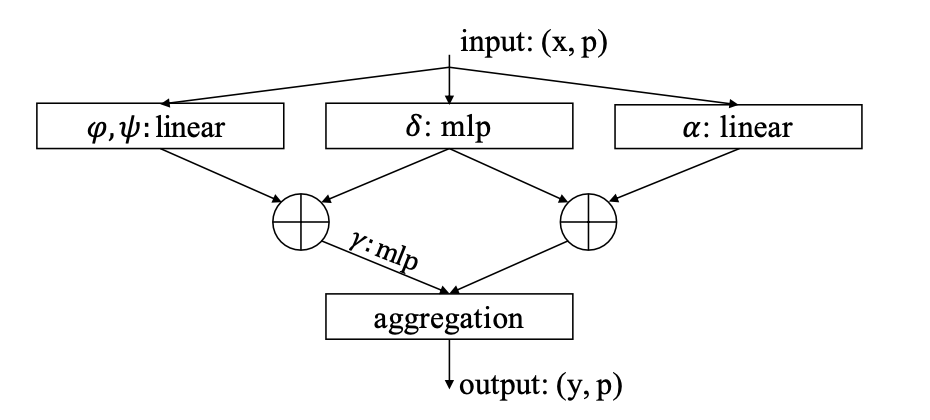}
\end{center}
   \caption{Point Transformer Layer Architecture \cite{zhao2021point}}
\label{fig:transformer_arch}
\end{figure}
% _______________________________________________________________
In figure \ref{fig:transformer_arch}, $\phi$, $\psi$, and $\alpha$ are pointwise feature transformations, such as linear projections and MLP's. $\delta$ is a position encoding function and $\rho$ is a normalization function such as softmax.\\

% _______________________________________________________________
\begin{figure}[hbt]
\begin{center}
   \includegraphics[width=0.8\linewidth]{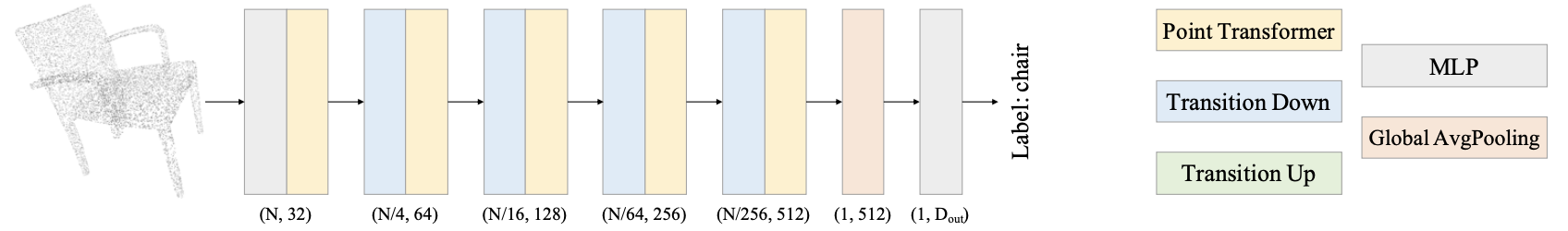}
\end{center}
   \caption{Point Cloud Classifier Architecture \cite{zhao2021point}}
\label{fig:classifier_arch}
\end{figure}
% _______________________________________________________________
In figure \ref{fig:classifier_arch}, the point transformer layer is combined with transition-down blocks.\\

We train our network on the ModelNet10 dataset, and then fine tune it for the 3D MNIST dataset. Additionally, we retrain the entire network on the 3D MNIST dataset and compare the results with those obtained from transfer learning.
% _______________________________________________________________
\begin{figure}[hbt]
\begin{center}
   \includegraphics[width=0.8\linewidth]{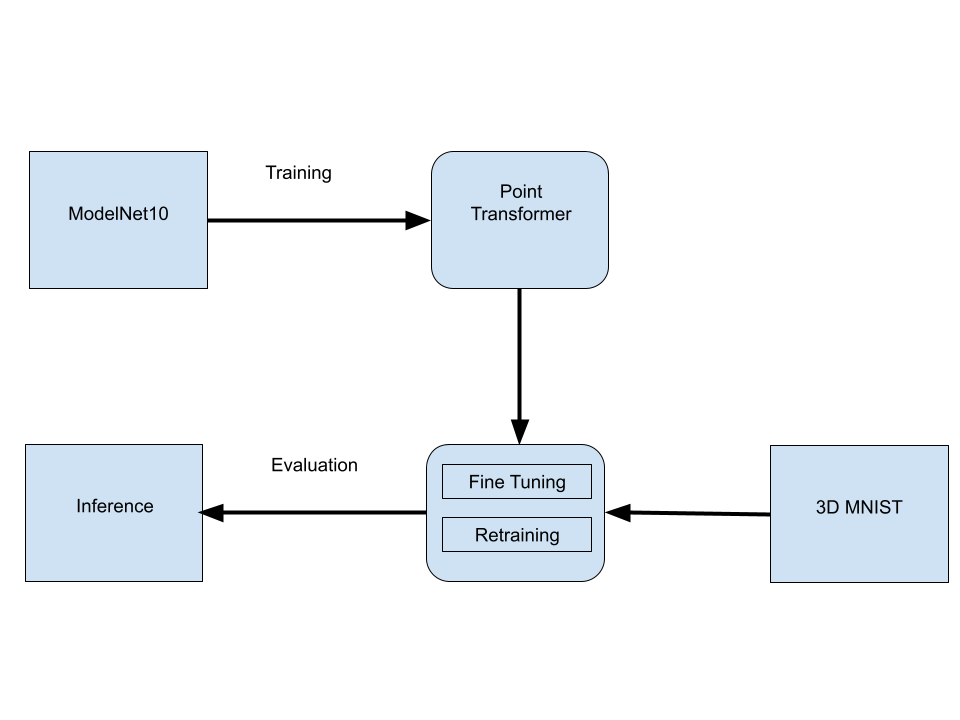}
\end{center}
   \caption{Methodology Employed}
\label{fig:method}
\end{figure}
% _______________________________________________________________

\section{Experimental Evaluation}
\begin{figure}[hbt]
\begin{center}
   \includegraphics[width=0.8\linewidth]{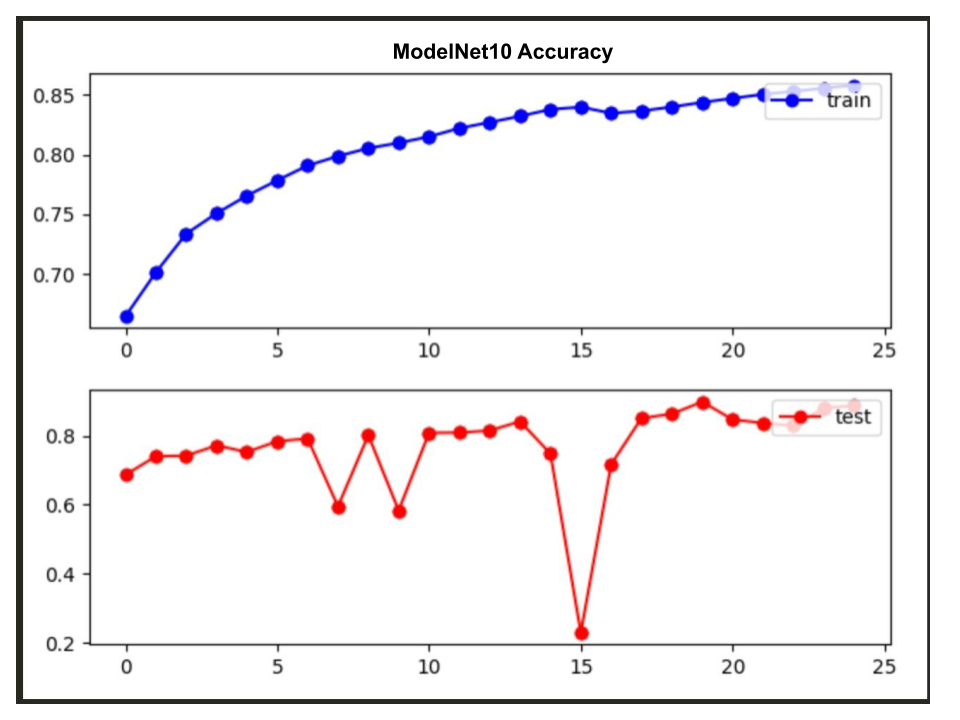}
\end{center}
   \caption{Training on ModelNet10}
\label{fig:training}
\end{figure}

We observed a training accuracy of 87.7\% while training the network on ModelNet10. Using the best model, we then apply fine tuning, a transfer learning technique, and evaluate the model performance on the 3D MNIST dataset. \\
Figure \ref{fig:training} depicts the accuracies obtained while training the Point Transformer model on the ModelNet10 dataset. As expected, the model learns features well on this dataset and resulting in an increase in accuracy overall.\\
% _______________________________________________________________
% _______________________________________________________________

\begin{figure}[hbt]
\begin{center}
   \includegraphics[width=0.8\linewidth]{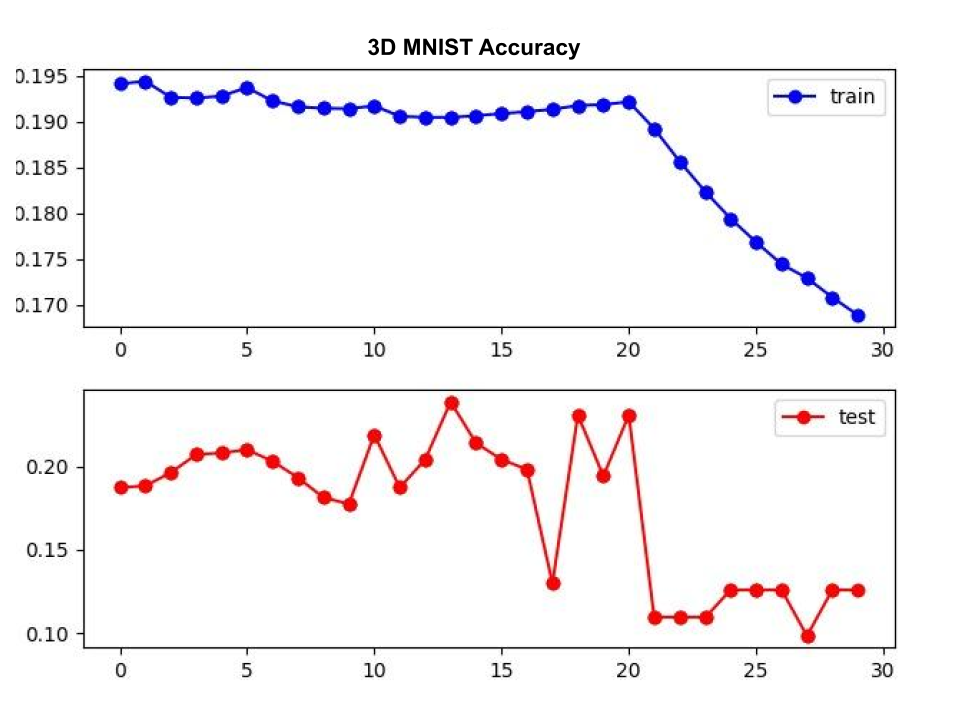}
\end{center}
   \caption{Training on 3D MNIST}
\label{fig:training_mnist}
\end{figure}

Figure \ref{fig:training_mnist} depicts the accuracies obtained while re-training the Point Transformer model on the 3D MNIST dataset. The network does not seem to learn the optimal parameters for 3D MNIST dataset. Although the expected accuracy in random initialisation should be 0.1\%, we are getting an accuracy of 0.25\%. This implies that the model seems to be learning some features but not enough to do provide a good classification performance.\\
% _______________________________________________________________
% _______________________________________________________________

\begin{figure}[hbt]
\begin{center}
   \includegraphics[width=0.8\linewidth]{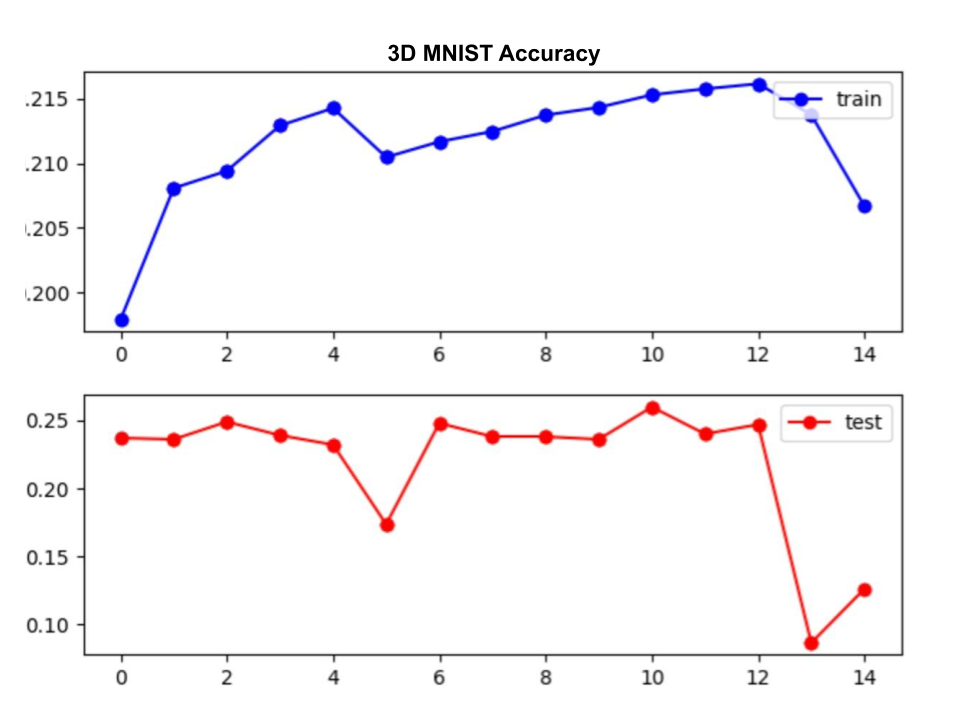}
\end{center}
   \caption{Fine Tuning on 3D MNIST}
\label{fig:tl_mnist}
\end{figure}

Figure \ref{fig:tl_mnist} depicts the accuracies obtained while fine tuning the ModelNet10 trained Point Transformer model on the 3D MNIST dataset for 15 epochs. We see that finetuning approach does not give good results for 3D MNIST dataset which can be explained by OOD shifts from ModelNet10 and 3D MNIST. However, the finetuned model converges faster than the pretrained one which shows that TL is still effective in learning lower level features such as edges and corners.
% _______________________________________________________________
% _______________________________________________________________

The Table \ref{tab:evaluation_metric} summarises the results obtained for 3D MNIST dataset. We observe that the model gives similar results for both retraining and fine tuning. However, fine tuning works for fewer epochs which indicates that transfer learning is effective to some extent.
\begin{table}[hbt]
\centering
\begin{tabular}{cc|cc|}
\cline{3-4}
\multicolumn{1}{l}{}         & \multicolumn{1}{l|}{} & \multicolumn{2}{c|}{Evaluation Metrics}  \\ \hline
\multicolumn{1}{|c|}{Epochs} & Method                & \multicolumn{1}{c|}{Accuracy} & F1 Score \\ \hline
\multicolumn{1}{|c|}{15}     & Fine Tuning           & \multicolumn{1}{c|}{26}       & 14.2     \\ \hline
\multicolumn{1}{|c|}{30}     & Retraining            & \multicolumn{1}{c|}{24.6}     & 11.6     \\ \hline
\end{tabular}
\label{tab:evaluation_metric}
\end{table}

We draw some inferences from the results tabulated in the Table \ref{tab:evaluation_metric}. Transfer learning relies on the assumption that the training data and the target data have similar underlying data distributions. Yet, if the out-of-distribution (OOD) data differs significantly from the source data's distribution, the knowledge transferred from the source may not be relevant or valuable. Consequently, the model may struggle to adapt and perform poorly when faced with the novel data distribution.

As the source dataset and the target dataset are too dissimilar, transfer learning was not effective. The model was not able to generalise the learned knowledge from ModelNet10 and apply it to classify 3D MNIST. Clearly, both the datasets require fundamentally different information for classification.\\

\begin{figure}[hbt]
\begin{center}
   \includegraphics[width=0.8\linewidth]{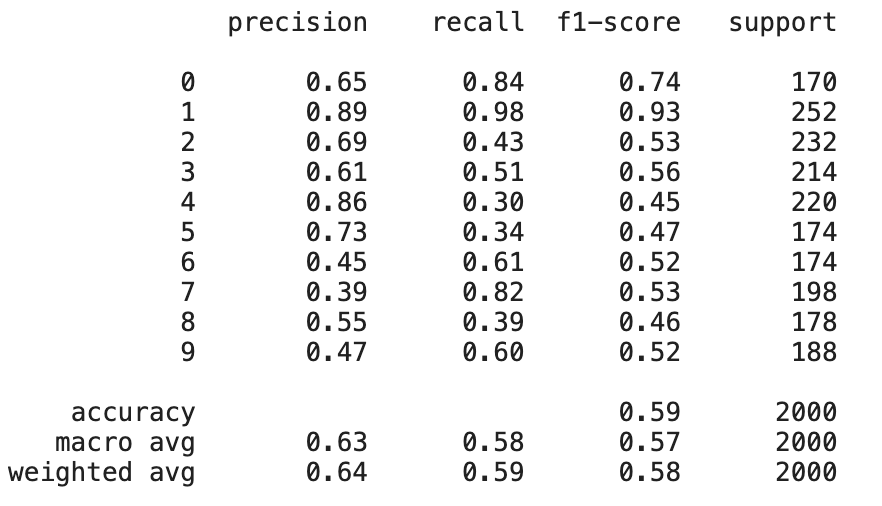}
\end{center}
   \caption{Training MLP model on 3D MNIST}
\label{fig:kaggle_results}
\end{figure}

To analyse classification on 3D MNIST, we create a new model specifically for 3D MNIST dataset which relies on basic MLP architecture. Figure \ref{fig:kaggle_results} tabulates the metrics obtained with the MLP-based model on 3D MNIST. Reverting to a simpler model made up of four-dense layers and two fully-connected layers gives better results. While inconclusive, attention-based mechanisms do not seem to learn the features from the 3D MNIST dataset. However, this needs to be explored further.
% _______________________________________________________________
% _______________________________________________________________

%-------------------------------------------------------------------------

{\small
\bibliographystyle{ieee_fullname}
\bibliography{main}
}

\end{document}